\documentclass[journal]{IEEEtran}

\usepackage{graphicx}
\usepackage{algorithm}
\usepackage{array}
\usepackage{amsmath}
\usepackage{url}
\usepackage{enumerate}
\usepackage{multicol}
\usepackage{textcomp}
\usepackage{multirow}
\usepackage{cite}
\usepackage{fmtcount}
\usepackage{diagbox}

\usepackage{color}
\definecolor{Change}{rgb}{0.5,0.1,0.1}

\usepackage{threeparttable}

\usepackage{changes}
\definechangesauthor[name={Per cusse}, color=orange]{per}
\usepackage{algorithm,algorithmicx,algpseudocode}

\usepackage{varwidth}

\begin{document}
	
	\title{Incorporating Human Domain Knowledge in 3D LiDAR-based Semantic Segmentation}
	
	\author{Jilin~Mei,~\IEEEmembership{Member,~IEEE,} Huijing~Zhao,~\IEEEmembership{Member,~IEEE}
		\thanks{This work is supported by the National Key Research and Development
				Program of China (2017YFB1002601) and the NSFC Grants (61573027).}
				
		\thanks{Correspondence: H. Zhao, zhaohj@cis.pku.edu.cn.}

		}

	\maketitle
	
	\begin{abstract}
	This work studies semantic segmentation using 3D LiDAR data. Popular deep learning methods applied for this task require a large number of manual annotations to train the parameters. We propose a new method that makes full use of the advantages of traditional methods and deep learning methods via incorporating human domain knowledge into the neural network model to reduce the demand for large numbers of manual annotations and improve the training efficiency. We first pretrain a model with autogenerated samples from a rule-based classifier so that human knowledge can be propagated into the network. Based on the pretrained model, only a small set of annotations is required for further fine-tuning. Quantitative experiments show that the pretrained model achieves better performance than random initialization in almost all cases; furthermore, our method can achieve similar performance with fewer manual annotations. 
	\end{abstract}

	
	\section{Introduction}
\begin{figure}
		\centering
		\includegraphics[width=0.5\textwidth]{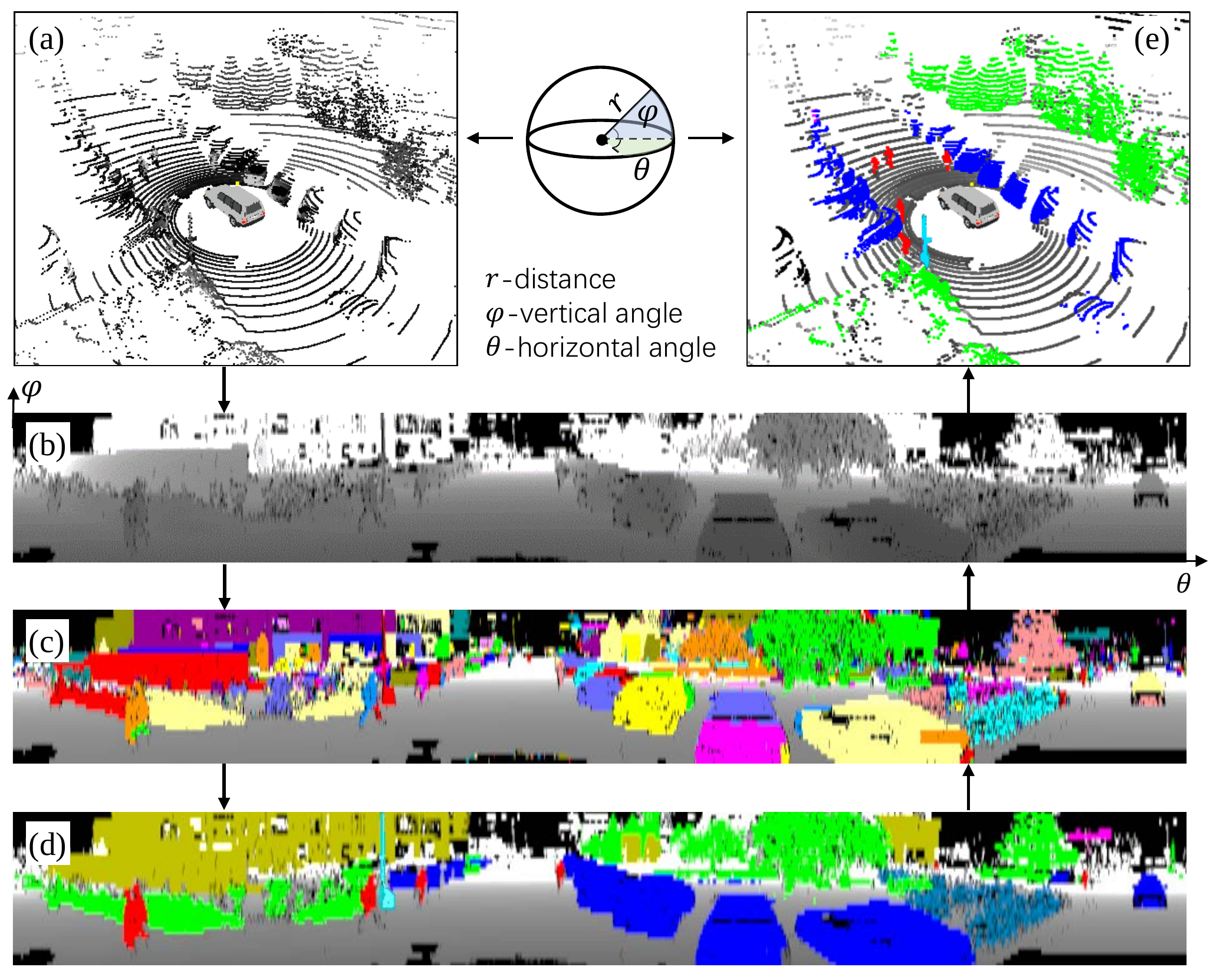}
		\caption{The 3D LiDAR-based semantic segmentation. (a) and (b) show the input data in two kinds of formats, i.e., 3D point cloud and 2D range frame. (c) is the result after over-segmentation. (d) and (e) show the semantic segmentation results.}
		\label{fig:intro1}
	\end{figure}

	Recently, 3D LiDAR sensors have been widely implemented as the ``eyes'' of autonomous driving systems\cite{urmson2008autonomous}. How to achieve efficient scene parsing, e.g., semantic segmentation, based on 3D LIDAR data has attracted increasing attention. Generally, semantic segmentation is the procedure of finding an object label for each point or data cluster\cite{munoz2009onboard}. In this work, the problem is defined as classification based on segmentation, as shown in Fig. \ref{fig:intro1}. Raw 3D LiDAR data can be equally represented by a range image (b) in the polar coordinate system, where the pixel value is the range distance. Then, oversegmentation is conducted on the range frame (c), and the problem of semantic segmentation is solved by discriminating the label of each segment (d).
	
	3D LiDAR-based semantic segmentation has been studied for the past decade\cite{munoz2009onboard,zhao2010scene,dewan17iros}. The traditional methods use handcrafted features\cite{zhao2010scene} that have a clear definition in the real world, e.g., the width is used to distinguish people from cars. Thus, these methods are interpretable. However, the adaptability of features for different scenes remains a challenge, and expert knowledge is required to adjust the parameters of the classifier.
	
	The outstanding performance of deep learning methods in image semantic segmentation\cite{garcia2017review} has encouraged researchers to apply these methods to 3D LiDAR data. These data-driven methods avoid handcrafted features by using abundant annotated data. However, the generation of fine annotations, especially for 3D LiDAR data, is challenging, and few public datasets are available for 3D LiDAR-based semantic segmentation aimed at autonomous driving applications.
	
	The two aforementioned types of methods have their own advantages and disadvantages. Traditional methods rely on human domain knowledge, while deep learning methods are data driven. Is there a way to combine the advantages and make up for the shortcomings? To the best of the author's knowledge, two techniques have been reported in the literature. The first is a semisupervised approach\cite{yan2006discriminative} that converts human prior knowledge into constraint information, such as data associations between frames, and adds the constraint item to the loss function. We have verified the effectiveness of this method in our previous work\cite{mei2018semantic}. The second is pretraining, e.g., initializing new networks with parameters trained on IMAGENET\cite{ILSVRC15}, which is widely used in tasks related to image processing. The effectiveness of pretraining is discussed in detail in \cite{Bengio2010}. However, it is unreasonable to initialize the network parameters directly from an image processing task, which motivates us to design a pretraining method suitable for 3D LiDAR data.

	In this paper, we make full use of the advantages of traditional methods and neural network methods via incorporating human domain knowledge into the neural network model to reduce the demand for large numbers of manual annotations and improve the training efficiency. The proposed method consists of two steps: parameter pretraining and parameter fine-tuning. In first step, a rule-based classifier based on human knowledge is designed, and samples are fed through the classifier to perform unsupervised classification. Then, these auto-annotated data are used to pretrain a convolutional neural network (CNN) to ensure that the parameters of the CNN fit the human knowledge. In the second step, the parameters in the pretrained CNN are transferred to a new network. Because the new CNN is well initialized, only a small number of manual annotations is necessary to update the parameters. We evaluate this method on a dynamic campus scene. Quantitative experiments show that the pretrained method has better performance than random initialization in almost all cases; furthermore, our method achieves similar performance with fewer manual annotations.
	
	The remainder of this paper is structured as follows. Sect. \uppercase\expandafter{\romannumeral2} discusses related work. The proposed method is presented in Sect. \uppercase\expandafter{\romannumeral3}. Sect. \uppercase\expandafter{\romannumeral4} shows the implementation details. Sect. \uppercase\expandafter{\romannumeral5} presents the experimental results. Finally, we draw conclusions in Sect. \uppercase\expandafter{\romannumeral6}.

	\section{Related Works}
	Semantic segmentation for 3D LiDAR data is not a new topic. We firstly review the methods on semantic segmentation, then discuss how to incorporate human knowledge. 
	\subsection{Semantic Segmentation}
	A popular method of semantic segmentation is classification of each point or data cluster. We separate the related literature into traditional methods and deep learning methods.
	
	In traditional methods, some researchers assume that each point or data cluster is independent; for example, \cite{weinmann2014semantic} presents a versatile framework including feature selection, feature extraction and classification. One-frame LiDAR data usually have millions of points, so the direct method is time consuming. Classification based on a segmentation framework is proposed in \cite{zhao2010scene}, where only the label of a cluster is evaluated. Some works consider the spatial relationship between elements, e.g., via the Markov random Field (MRF) or conditional random field (CRF). Features are embedded in node potentials, and spatial relationships are encoded in edge potentials \cite{munoz2009onboard}. The solution of a CRF or MRF sometimes requires high-dimensional optimization. Thus, \cite{lu2012simplified} proposes a simplified MRF, where the node and edge potentials are directly updated, and \cite{wang2015efficient} attempts to simplify the point clouds via voxel-neighbor structure. The main disadvantage of traditional methods is the adaptability of handcrafted features to different scenes.
	
	In deep learning methods, features are automatically learned from data. Researchers focus mainly on discussing data representations and new network structures. Inspired by image semantic segmentation, raw 3D data can be converted into 2D images. \cite{tosteberg2017semantic} uses virtual 2D RGB images obtaine via Katz projection, and \cite{dewan17iros,wu2017squeezeseg} unwrap  3D LiDAR data on spherical range image. Another stream of research considers 3D representations. Voxel occupancy grid is a good way to make irregular raw LiDAR data grid-aligned \cite{hackel2017isprs}. The voxel representation is further improved in OctNet\cite{riegler2017octnet}, which has more efficient memory allocation and computation. New network structures specified for 3D data are also studied. PointNet\cite{qi2017pointnet} directly takes raw point clouds as input, and a novel type of neural network is designed with multilayer perceptrons (MLPs). Based on \cite{qi2017pointnet}, \cite{engelmann2017exploring} extends the method to incorporate larger-scale spatial context, and \cite{landrieu2017large} proposes the superpoint graph to capture the contextual relationships from point clouds. The main shortcoming of deep learning methods is the demand for large amounts of manual annotations.
	
	\subsection{Incorporating Human Knowledge}
	The purpose of incorporating human knowledge is to reduce the need for large numbers of manual annotations in deep learning methods. To the best of the author's knowledge, two types of methods have been reported in the literature: semi/weakly-supervised learning and model pretraining. 
	
	Semi/weakly-supervised learning involves introducing a few fine annotations or large numbers of ambiguous annotations during parameter learning. \cite{bearman2016s} uses point supervision, where annotators are asked to point to an object if one exists. Then, the point annotation and objectness prior are incorporated into the loss function to train the neural network. Similar work is proposed in ScribbleSup\cite{lin2016scribblesup}, which trains convolutional networks for semantic segmentation supervised by scribbles. The size constraint of an object is considered in \cite{papandreou2015weakly}. Our previous work on semisupervised learning implements a pairwise constraint, which encourages associated samples to be assigned the same labels\cite{mei2018semantic}. Some works focus on training models in weakly supervised settings, such as image-level tags\cite{pathak2015constrained}, bounding box labels\cite{dai2015boxsup} and unlabeled examples\cite{xu2015learning}; these weak supervision methods can be applied separately or in combination.
	
	During model pretraining, the target network is initialized with the parameters from a pretrained baseline network, e.g., the parameters in the first n layers copy from the baseline network and the remaining parameters are set randomly. This strategy has been widely used in image processing tasks, e.g., segmentation\cite{long2015fully} and transfer learning\cite{Yosinski2014}. The baseline network is usually trained with IMAGENET\cite{ILSVRC15}. The experiments in \cite{Bengio2010} confirm and clarify the advantages of unsupervised pretraining, which provides a good initial marginal distribution and exhibits properties of a regularizer. \cite{wulfmeier2016incorporating} introduces human priors by pretraining a model to regress a cost function under the framework of inverse reinforcement learning. Inspired by \cite{wulfmeier2016incorporating}, we propose a pretraining method suitable for 3D LiDAR data in which large numbers of auto-annotated data are generated with human domain knowledge.
	
	\begin{figure*}
		\centering
		\includegraphics[width=0.75\textwidth]{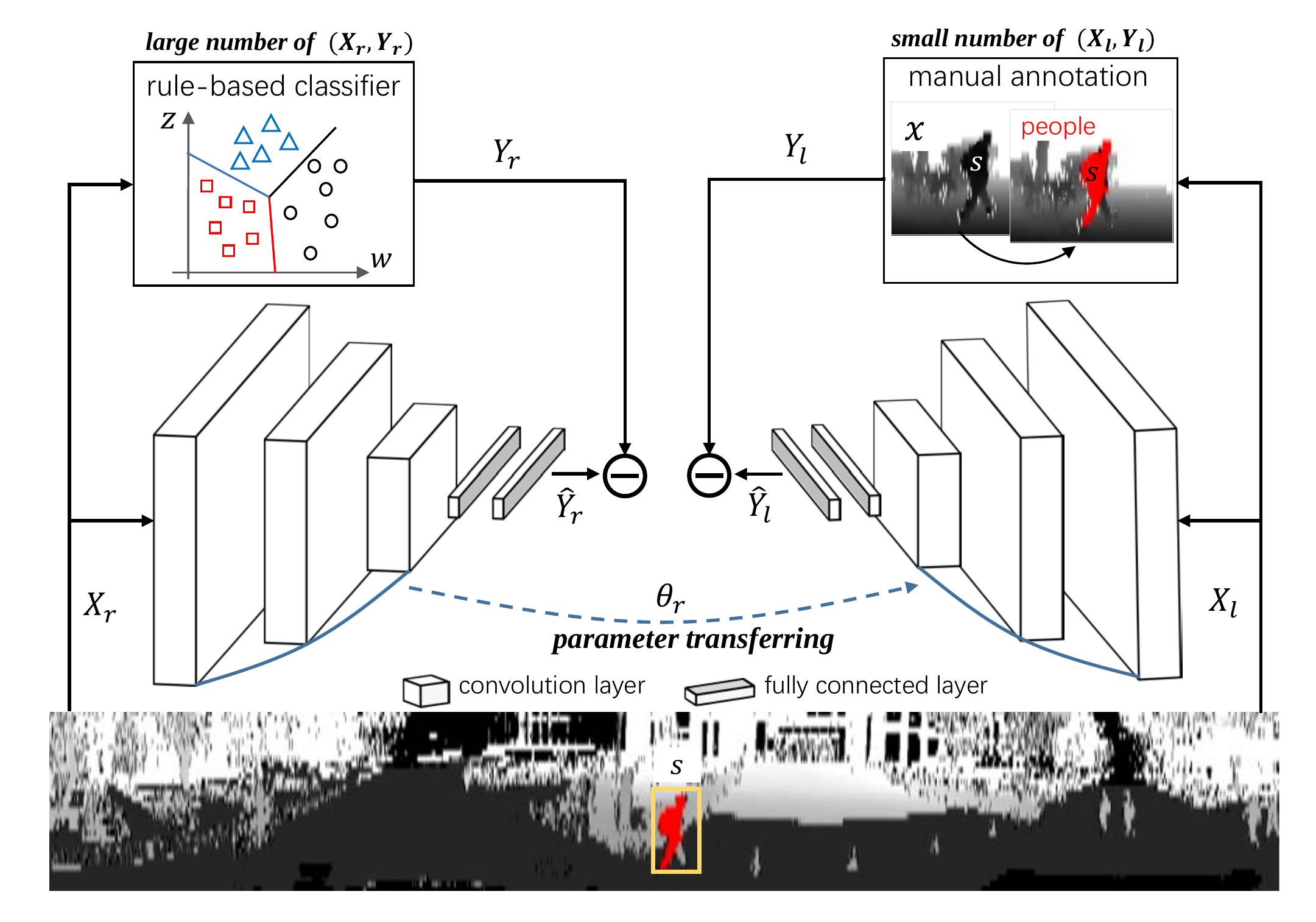}
		\caption{The framework of incorporating human knowledge for 3D LiDAR-based semantic segmentation. The left part is pretraining step and the right is fine-tuning.}
		\label{fig:workflow}
	\end{figure*} 
	
	\begin{figure}
		\centering
		\includegraphics[width=0.5\textwidth]{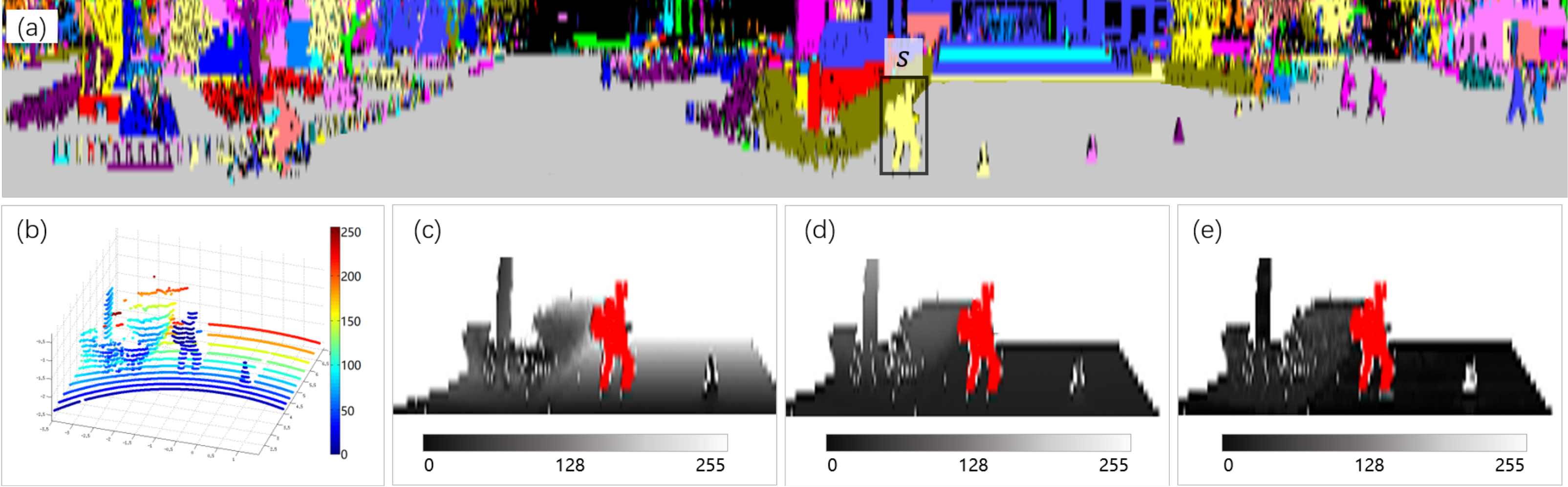}
		\caption{The procedure of sample generation from segment. (a) the segment $s$ is chosen as candidate region. (b) the neighbor points around $s$ are cropped to make one sample which has three channels. (c) the range channel, and we mark $s$ with red for better visualization. (d) the height channel. (e) the intensity channel. Please refer to \cite{mei2018semantic} for details.}
		\label{fig:samplegeneration}
	\end{figure} 	
		
	\section{METHODOLOGY}
	\subsection{Problem Definition}
	\label{3_A}
	 Let $P$ be the range frame converted from raw 3D point clouds. Segments $\{s_i\}_{i=1}^{N}$ are obtained on $P$ by evaluating the similarity of 3D points with their neighborhoods, e.g., a region growing method. We assume that one segment $s$ measures only a single object after oversegmentation. As $s$ commonly represents a part of the object, a data sample $x$, including $s$ and the surrounding data, is defined at the center of $s$, as shown in Fig.\ref{fig:workflow}. In one range frame $P$, $\{s_i\}_{i=1}^{N}$ and $\{x_i\}_{i=1}^{N}$ can be equally converted into each other.
	 The problem in this work is formulated as learning a multiclass classifier $f_\theta$ that maps $x$ to a label $y \in \{1,...,K\}$ and subsequently associates $y$ with the 3D points of $s$.
	
	\begin{equation}
	f_{\theta}: x\to y \in \{1,...,K\}
	\end{equation}
	
	Given a set of annotated data $X=\{x_i\}_{i=1}^{M},Y=\{y_i\}_{i=1}^{M}$, where $\{y_i\}$ is a one-hot label for $\{x_i\}$, a common way of learning a classifier $f_\theta$ is to find the best $\theta^*$ that minimizes a loss function $L$, i.e., the cross entropy, as below.
	
	\begin{equation}
	\label{equation:loss_func}
	\begin{split}
	&\theta^{*}=\mathop{\arg\min}_{\theta}L(\theta;X,Y) \\
	&L(\theta;X,Y) = -\frac{1}{M}\sum_{i=1}^M\sum_{k=1}^K{\textbf{1}[y^{k}_{i}=1]ln(P^k_{\theta}(x_i))},
	\end{split}
	\end{equation}
	 where $\textbf{1}[*]$ is an indicator function and $P^k_{\theta}(x_i)$ is the probability that $x_i$ is assigned a label $k$ by a classifier with parameters $\theta$.
	Stochastic gradient descent (SGD) is commonly applied to solve $\theta^{*}$. Thus, Equation (\ref{equation:loss_func}) can be rewritten as:
	
	\begin{equation}
		\theta^{*}=\mathop{\arg\min}_{\theta}L(\theta;X,Y, \theta_{0}),
	\end{equation}	
	where $\theta_{0}$ is the start position of SGD. $\theta_{0}$ can be obtained from a random distribution, e.g., a truncated normal distribution; or initialized from human knowledge via pretraining, such as in the proposed method.
	
	\subsection{Work Flow}
	As shown in Fig. \ref{fig:workflow}, the framework consists of two steps: parameter pretraining and parameter fine-tuning. One range frame is divided into multiple segments, and each segment corresponds to one sample $x$ in Fig. \ref{fig:samplegeneration}. Consequently, we can automatically produce a large number of samples. The sample generation follows that of \cite{mei2018semantic}. 
	
	During the pretraining step, we design a rule-based classifier that incorporates human knowledge. The unlabeled samples $X_r$ are passed through this classifier to predict the label; thus, the auto-annotated data $(X_r,Y_r)$ are obtained. The rule-based classifier works in an unsupervised manner, and we need not train the classifier via manual annotations. Combining $X_r$ and $Y_r$, a CNN is pretrained with a random initialization $\theta_0$, and the parameters are updated via back-propagation. 
	\begin{equation}
		\theta_r=\mathop{\arg\min}_{\theta}L(\theta;X_r,Y_r, \widehat{Y}_r, \theta_{0}),
	\end{equation}	
	where $\widehat{Y}_r$ is the output of the CNN. In this way, we obtain the pretrained parameter $\theta_r$. $\widehat{Y}_r$ will continue to fit $Y_r$, so we assume that the human rules can be propagated into the CNN.
	
	In the fine-tuning step, the pretrained parameters $\theta_r$ are first transferred into a new neural network. Since the new classifier is well initialized, we need only a small number of manual annotations $(X_l,Y_l)$ for parameter fine-tuning. The parameter updating is the same as in the first step, except that the start position of the optimization is different. The final classifier is obtained by:
	
	\begin{equation}
	\theta^{*}=\mathop{\arg\min}_{\theta}L(\theta;X_l,Y_l, \widehat{Y}_l, \theta_{r}),
	\end{equation}
	where $\widehat{Y}_l$ is the output of the CNN and $\theta_{r}$ is the start position.
			
	\subsection{Rule-based Classifier}  
	
	\begin{figure}
		\centering
		\includegraphics[width=0.4\textwidth]{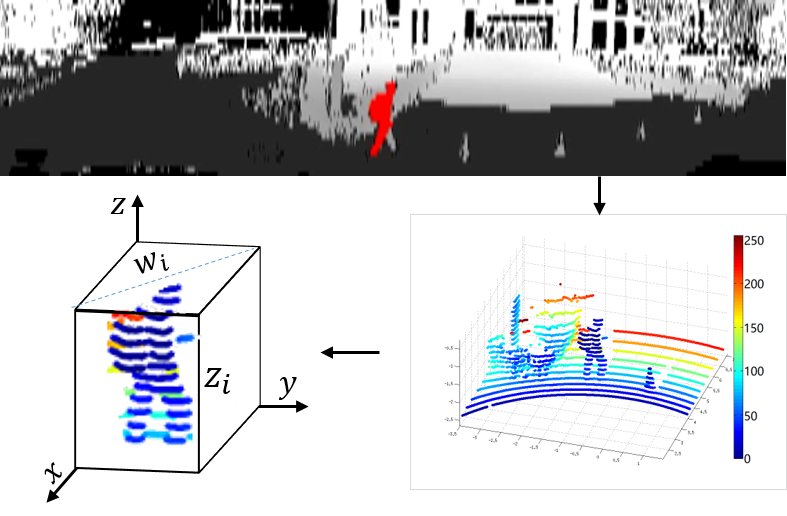}
		\caption{The definition of features for rule-based classifier. The height $z_i$ and width feature $w_i$ of segment $s_i$ are evaluated.}
		\label{fig:rules}
	\end{figure}             
	
	LiDAR can directly measure distance information without being affected by illumination, and this robust attribute motivates us design a classifier based on rules in the real world. The sample and segment are associated as detailed in \ref{3_A}; thus, only features of the segment are considered for each sample in the rule-based classifier. As shown in Fig. \ref{fig:rules}, the height and width of a segment are calculated from the raw point cloud, and these two features reflect the physical attributes of objects in the real world, i.g., the height of a car generally does not exceed 2 m. We believe that the width and height of objects are efficient information to design a simple classifier. As shown in Algorithm \ref{rule-based}, a trunk is higher than 2 m and has a width in (0, 2.5], people are shorter than 2 m but have widths larger than 0.2 m, a car is shorter than 2 m and has a width in [1.5,2.5], and so on. These rules can easily be understood by a human, however, if we want the CNN to understand these rules, the general approach is to train the CNN with a large number of manual annotations. The proposed rule-based classifier can automatically generate annotations without human effort, which accelerates the CNN learning.

	\begin{algorithm}
		\caption{\small Rule-based Classifier}
		\label{rule-based}
		\begin{algorithmic}[1]
			\Require all segments $S$ in one range frame
			\Ensure labeling results $\Phi$
			\State Initialize $\Phi$ with $\emptyset$
			\ForAll{$s_{i}$ \textbf{in}  $S$}
			\State $label=Unknown$
			\State calculate the width $w_i$ and height $z_i$ of $s_{i}$
			\If {$w_i\in$ [0,2.5] \textbf{and} $z_i>2.0$}
			\State $label=Trunk$	\Comment{Trunk is slim and high }
			\ElsIf {$w_i\in$ [0,1.5]}
			\If {$w_i>0.2$}
			\State $label=People$ \Comment{People is shorter than 2m}
			\EndIf
			\ElsIf {$w_i\in$ [1.5,2.5]}
			\If {$z_i<2.0$}
			\State {$label=Car$} \Comment{Car is wider than People}
			\EndIf		
			\ElsIf {$w_i\in$ [8.0,15]}
			\State $label=Building$ \Comment{Building is flat}
			\EndIf	
			\State $\Phi\leftarrow<s_i,label>$	
			\EndFor					
		\end{algorithmic}
	\end{algorithm}

	\begin{table*}
		\centering
		\begin{threeparttable}
			\renewcommand\arraystretch{1.5}			
			\small
			\caption{The samples of manual annotations.}
			\label{tab:dataset}
			
			\begin{tabular}{|c|c|c|c|c|c|c|c|}
				\hline
				\diagbox[width=5em,height=1.3em] & people & car  & trunk & bush & building & cyclist & unknown \\ \hline
				training set & 1533   & 6014 & 1837  & 9064 & 7736     & 366     & 5113    \\ \hline
				testing set  & 1880   & 5074 & 1746  & 9102 & 3230     & 562     & 4630    \\ \hline
			\end{tabular}
			
		\end{threeparttable}
	\end{table*}	
	
	\begin{table*}
		\centering
		\begin{threeparttable}
			\renewcommand\arraystretch{1.5}			
			\small
			\caption{The performance of rule-based classifier.}
			\label{tab:rule-based}
			
			\begin{tabular}{|c|c|c|c|c|c|}
				\hline
				\diagbox[width=5em,height=1.2em]& people & car  & trunk & building & unknown \\ \hline
				labeling on training set & 3009   & 6255 & 5914  & 3596     & 9723    \\ \hline
				F1 score on testing set & 50.0   & 64.8 & 48.9  & 62.6     & 66.0    \\ \hline
			\end{tabular}
			
		\end{threeparttable}
	\end{table*}

	\subsection{Parameter Pretraining}

	Parameter pretraining has been widely used in image processing tasks such as classification, detection and segmentation, but it is unreasonable to initialize a network toward LiDAR data using the parameters from image processing tasks, as these two types of data are substantially different in terms of both human visual and physical meaning. Therefore, a pretraining approach should be designed for LiDAR data. Now, the question is why pretraining reduces the need for manual annotations during the training phase of neural networks? We answer the question from two perspectives.
	
	\textbf{Perspective of probability.} Human knowledge describes the distribution of the input data $X$, i.e., $P(X)$, which represents data priors, and the target CNN classifier can be treated as a conditional probability that predicts the label $Y$ for each input data, i.e., $P(Y|X)$. In general, the CNN is trained with only human annotations that are similar to the joint distribution $P(X,Y)$, but data priors $P(X)$ are ignored. Based on Bayes rule, $P(Y|X)=P(Y,X)/P(X)$, if accurate priors are supported, we believe that the dependence on manual annotation can be reduced. We design a rule-based classifier to obtain $P(X)$ from human domain knowledge.
			
	\textbf{Perspective of gradient descent.} Gradient descent is the conventional parameter updating method of CNNs. The selection of the initial position largely determines whether gradient descent can converge to the global minimum. Random initialization is a common strategy. The initial position is randomly selected in the high-dimensional parameter space, which increases the possibility of training results falling into local minima. In our method, the samples used for pretraining are supervised by human rules, that is, autogenerated from the rule-based classifier. Thus, the network will continue to fit these rules, and the performance of the pretraining network is related to the rule-based classifier. Although we cannot assume that the pretrained parameters are optimal, they are reasonable. When the initial position of gradient descent starts from the pretrained parameters, the dependence on manual annotation is reduced.

	\begin{figure}
		\centering
		\includegraphics[width=0.4\textwidth]{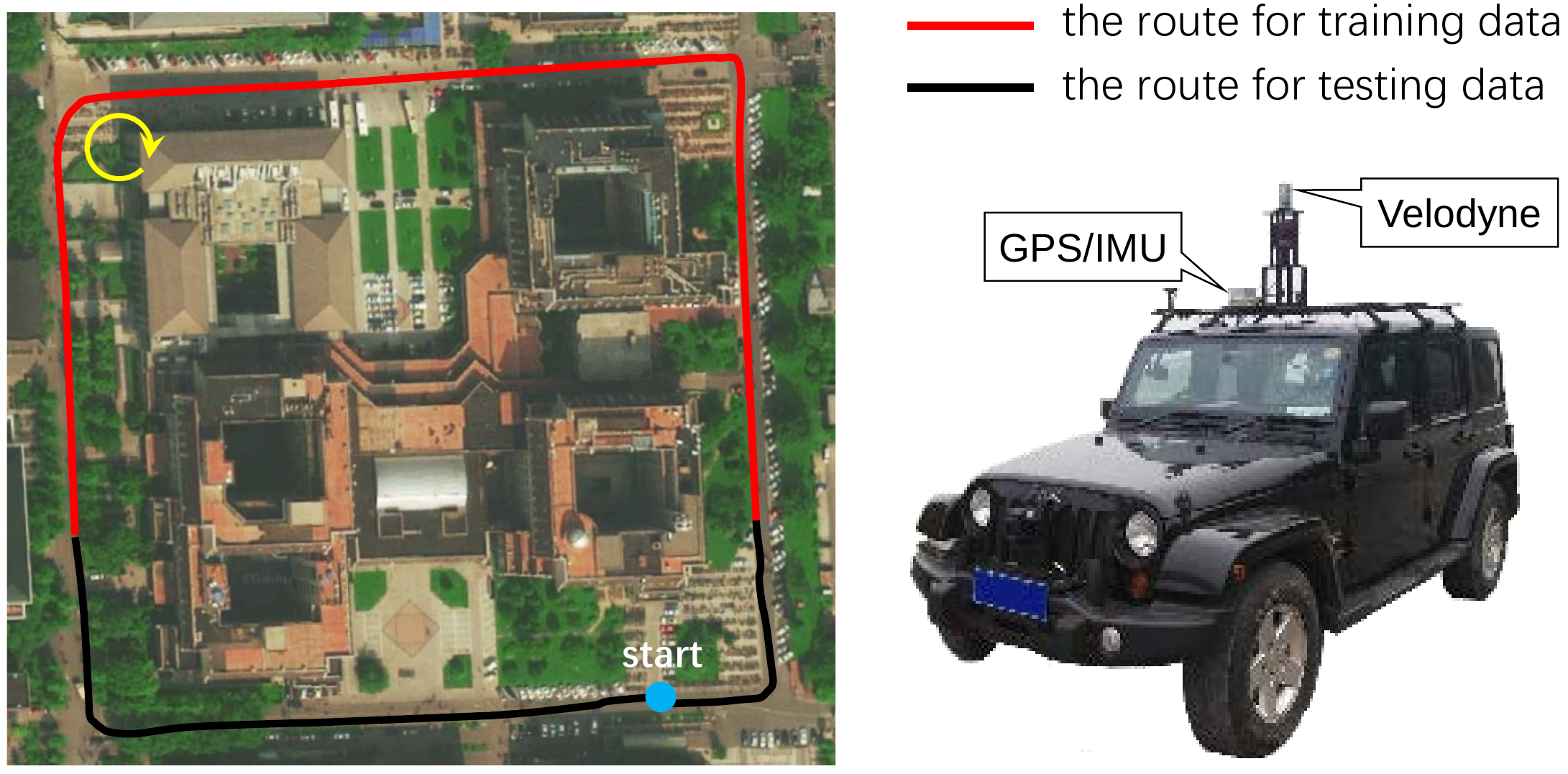}
		\caption{The routes of data collection and the platform configuration.}
		\label{fig:data_route}
	\end{figure}	
		
	In our method, the samples for pretraining are autogenerated by the rule-based classifier, and we do not require any human annotation in this step.
		
	\subsection{Parameter Fine-tuning}
	\label{sect:param-finetune}
	We define parameter fine-tuning as initializing a new network with pretrained parameters and training the network with manual annotations. The main body of a CNN consists of convolutional layers and fully connected (FC) layers. Generally, there are two ways to perform parameter fine-tuning in the context of a CNN. Both of them copy the parameters of convolutional layers from a pretrained network to the new network and randomly set the FC layers. The difference is whether the paramters in th convolutional layers are fixed during fine-tuning. We test these two configurations in experiments.

	\section{Implementation Details}
	
	\label{sect:implement-details}
	The CNN used here consists of three convolutional layers whose dimensions in [width,height,depth] are [256,256,32], [128,128,32], and [64,64,64]; two fully connected layers whose dimensions are both [128,1]; and one softmax layer. In parameter fine-tuning, the CNN predicts 7 labels: people, car, trunk, bush, building, cyclist and unknown. In pretraining, the CNN predicts only 5 labels, namely, people, car, trunk, building and unknown since the rule-based classifier cannot fully discriminates cyclist and bush.
	
	All networks are trained under the TensorFlow framework using ADAM solver and a learning rate of 1e-4. The batch size is 2, and we save a checkpoint every 100 iterations. The training phase stops when the loss converges or is less than 1e-4. For each classifier, we evaluate all checkpoints on the testing set and select the checkpoint with the highest F1 score. 

	\begin{table*}
		\centering
		\begin{threeparttable}
			\renewcommand\arraystretch{1.5}			
			\small
			\caption{The comparisons with F1 measure on testing set.}
			\label{tab:pretrained}
			
			\begin{tabular}{|c|c|c|c|c|c|c|}
				\hline
				classifier & people & car  & trunk & building & unknown & mean score \\ \hline
				rule-based     & 50.0   & 64.8 & 48.9  & 62.6     & 66.0    & 58.5       \\ \hline
				pretrained-CNN & 57.4   & 67.0 & 50.4  & 66.0     & 67.2    & 61.6       \\ \hline
			\end{tabular}
			
		\end{threeparttable}
	\end{table*}

	\section{EXPERIMENTAL RESULTS}

	\begin{figure}
		\centering
		\includegraphics[width=0.45\textwidth]{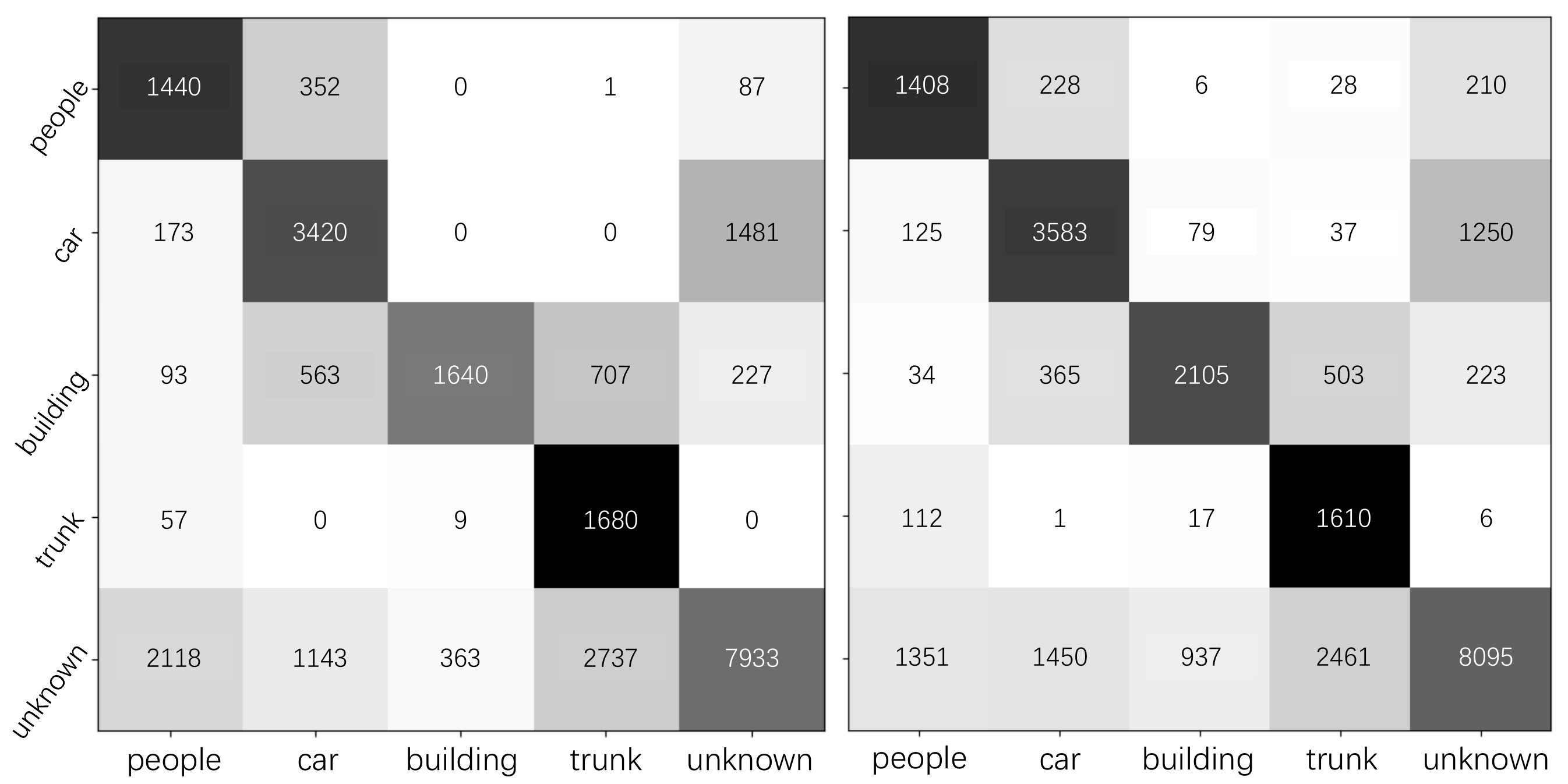}
		\caption{The comparison between rule-based and CNN-pretrained classifier. The color is indexed by recall, and a darker color means a higher recall. (a) The confusion matrix of rule-based classifier. (b) The confusion matrix of CNN-pretrained classifier.}
		\label{fig:cm-rule-based}
	\end{figure}

	\subsection{Data Set}
	The performance of the proposed method is evaluated on a
	dynamic campus dataset collected by an instrumented vehicle with a GPS/IMU suite and a Velodyne-HDL32, as shown in Fig. 7. The total route is approximately 890 meters. All sensor data are collected, and each data frame is associated with a time log for synchronization. The GPS/IMU data are logged at 100 Hz. The LiDAR data are recorded at 10 Hz and include 1039 frames of training data (red line in Fig. \ref{fig:data_route}) and 790 frames of testing data (black line in Fig. \ref{fig:data_route}). One frame can produce multiple samples; for example, we obtain 6014 car samples from the 1039 frames of training data in TABLE \ref{tab:dataset}.
	
	To make quantitative comparisons, manual annotations\cite{mei2018semantic} are conducted on both training and testing sets; the labeling results are shown in TABLE \ref{tab:dataset}.

	\begin{table*}
		\centering
		\begin{threeparttable}
			\renewcommand\arraystretch{1.5}			
			\small
			\caption{The subdivision of training set for funetuning.}
			\label{tab:datset-sub}
			
			\begin{tabular}{|c|c|c|c|c|c|c|c|}
				\hline
				finetuning data & people & car  & trunk & bush & building & cyclist & unknown \\ \hline
				training set & 1533   & 6014 & 1837  & 9064 & 7736     & 366     & 5113    \\ \hline
				sub-100      & 100    & 100  & 100   & 100  & 100      & 100     & 100     \\ \hline
				sub-400      & 400    & 400  & 400   & 400  & 400      & 366     & 400     \\ \hline
				sub-1600     & 1533   & 1600 & 1600  & 1600 & 1600     & 366     & 1600    \\ \hline
			\end{tabular}
			
		\end{threeparttable}
	\end{table*}
	
	\begin{table*}
		\centering
		\begin{threeparttable}
			\renewcommand\arraystretch{1.5}			
			\small
			\caption{The F1 score of different classifiers on testing set.}
			\label{tab:all-result}
			
			\begin{tabular}{|c|c|c|c|c|c|c|c|c|c|}
				\hline
				\multicolumn{1}{|l|}{finetuning data} 	& classifier    & people        & car           & trunk         & bush          & building      & cyclist       & unknown       & mean score          \\ \hline
				\multirow{2}{*}{sub-100}            	& baseline-100  & 46.0          & 56.5          & 66.3          & 65.5          & 56.0          & 30.8          & 35.8          & \textit{51.0}       \\
				& pretrain-100  & 55.0          & 71.6          & 68.6          & 63.9          & 61.8          & 31.3          & 39.4          & \textit{55.9}       \\ \hline
				\multirow{2}{*}{sub-400}            	& baseline-400  & 59.4          & 72.2          & 73.5          & 71.4          & 71.8          & 44.0          & 35.1          & 61.1                \\
				& pretrain-400  & 67.1          & 79.7          & 71.9          & 72.8          & 70.8          & 45.5          & 48.5          & \underline{65.2}          \\ \hline
				\multirow{2}{*}{sub-1600}           	& baseline-1600 & 68.7          & 80.3          & 74.4          & 70.8          & 69.9          & 46.1          & 48.9          & \underline{65.6}          \\
				& pretrain-1600 & 71.2          & 82.5          & 75.9 			& 75.9          & 72.6          &\textbf{46.5}           & 49.0          & \textit{\underline{67.7}} \\ \hline
				\multirow{2}{*}{training set}       	& baseline-all  & 69.2          & 85.1          & 77.2          & 75.3          & 76.8          & 44.6          & \textbf{53.8} & \textit{\underline{68.8}} \\
				& pretrain-all  & \textbf{71.6} & \textbf{87.5} & \textbf{80.2}          & \textbf{77.1} & \textbf{78.2} &45.4  & 53.6          & \textbf{70.5}       \\ \hline
			\end{tabular}
			
			\begin{tablenotes}
				\footnotesize
				\item[1] baseline-* : random initialization; pretrain-* : pretraining initialization(ours).
			\end{tablenotes}
			
		\end{threeparttable}
	\end{table*}
	
	\subsection{Rule-based Classifier}
	The rule-based classifier is simple, and its effectiveness should be assessed before conducting further experiments. As shown in TABLE. \ref{tab:rule-based}, we pass both training and testing sets through this classifier which assigns one label for each sample. Currently, the classifier supports only 5 categories. In this way, we collect the auto-annotated training set (the second row of TABLE. \ref{tab:rule-based}) for parameter pretraining. At the same time, the effectiveness is evaluated on the testing set in the term of F1 measure, which is defined as:
		\begin{equation}
		\label{eq:f-measure}
		F1-Measure = \frac{2*recall*precision}{recall+precision}\cdot{100}.
		\end{equation} 
	We merge bush and cyclist in TABLE. \ref{tab:dataset} into unknown when calculating the F1 score for the rule-based classifier, and the results are shown in the third row of TABLE. \ref{tab:rule-based}. 
	
	The F1 scores in TABLE. \ref{tab:rule-based} are encouraging: simple rules ensure reasonable results, and the rule-based classifier is effective. The confusion matrix is shown in Fig. \ref{fig:cm-rule-based} (a), one notable	result is that the recall of trunk is very high.

	\begin{figure}
		\centering
		\includegraphics[width=0.5\textwidth]{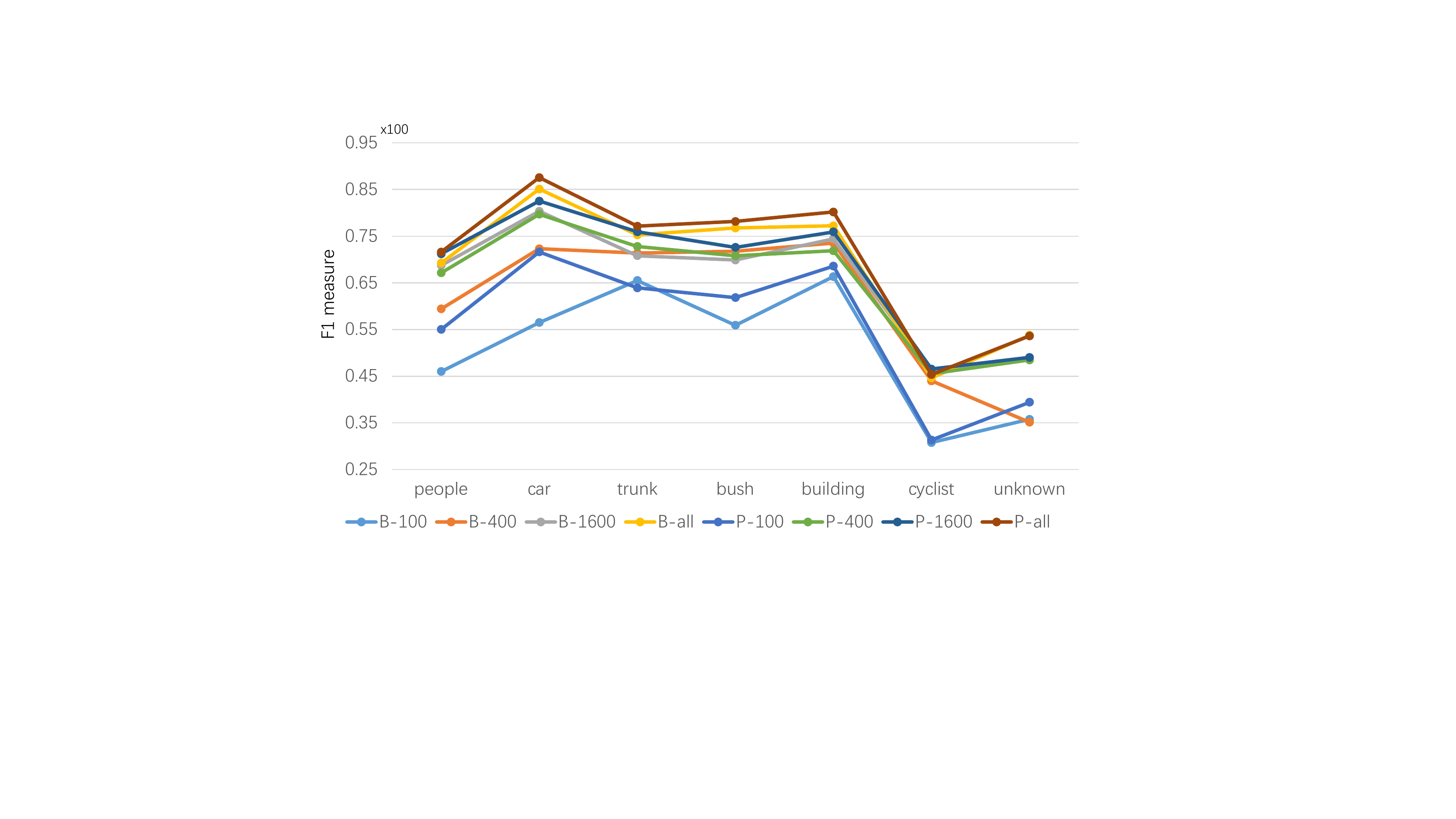}
		\caption{The quantitative comparison of different classifiers. B-* means baseline and P-* means pretrained classifier.}
		\label{fig:f1onclassifier}
	\end{figure}

	\subsection{Pretraining Results}
	The CNN is trained with the autolabeled samples in TABLE. \ref{tab:rule-based}. We expect the pretrained CNN to have similar performance to the rule-based classifier since the samples are supervised by the rules. As illustrated in TABLE. \ref{tab:pretrained}, the pretrained CNN has better F1 scores on each category than does the rule-based method. A more detailed comparison is in Fig. \ref{fig:cm-rule-based}. These two classifier has very similar performances, i.g., they both achieve high recalls on trunk. On the basis of these results, we assume that the rules designed by humans have propagated into the pretrained CNN and that the parameters of CNN are more reasonable than random initialization.
	
	We emphasize that pretraining dose not require any manual labeling. Although the samples are supervised by rules, the pretrained CNN still performs better than the rule-based classifier.
	
	\subsection{Fine-tuning Results}
	The fine-tuning data has four components as shown in TABLE. \ref{tab:datset-sub}. Sub-100$\sim$1600 are randomly selected from the raw training set. For example, sub-1600 means choosing 1600 samples from each category. If the number of sample is less than 1600, such as for people and cyclist categories, we do not perform any sample augmentation.
	
	As illustrated in TABLE. \ref{tab:all-result} and Fig. \ref{fig:f1onclassifier}, different classifiers are tested based on the fine-tuning data. Baseline-* means the parameters are initialized with the truncated normal distribution, and pretrain-* means the parameters in the convolutional layers copy from the pretrained network. All classifiers share the same network structure detailed in Sect. \ref{sect:implement-details}.
	
\begin{figure}
		\centering
		\includegraphics[width=0.5\textwidth]{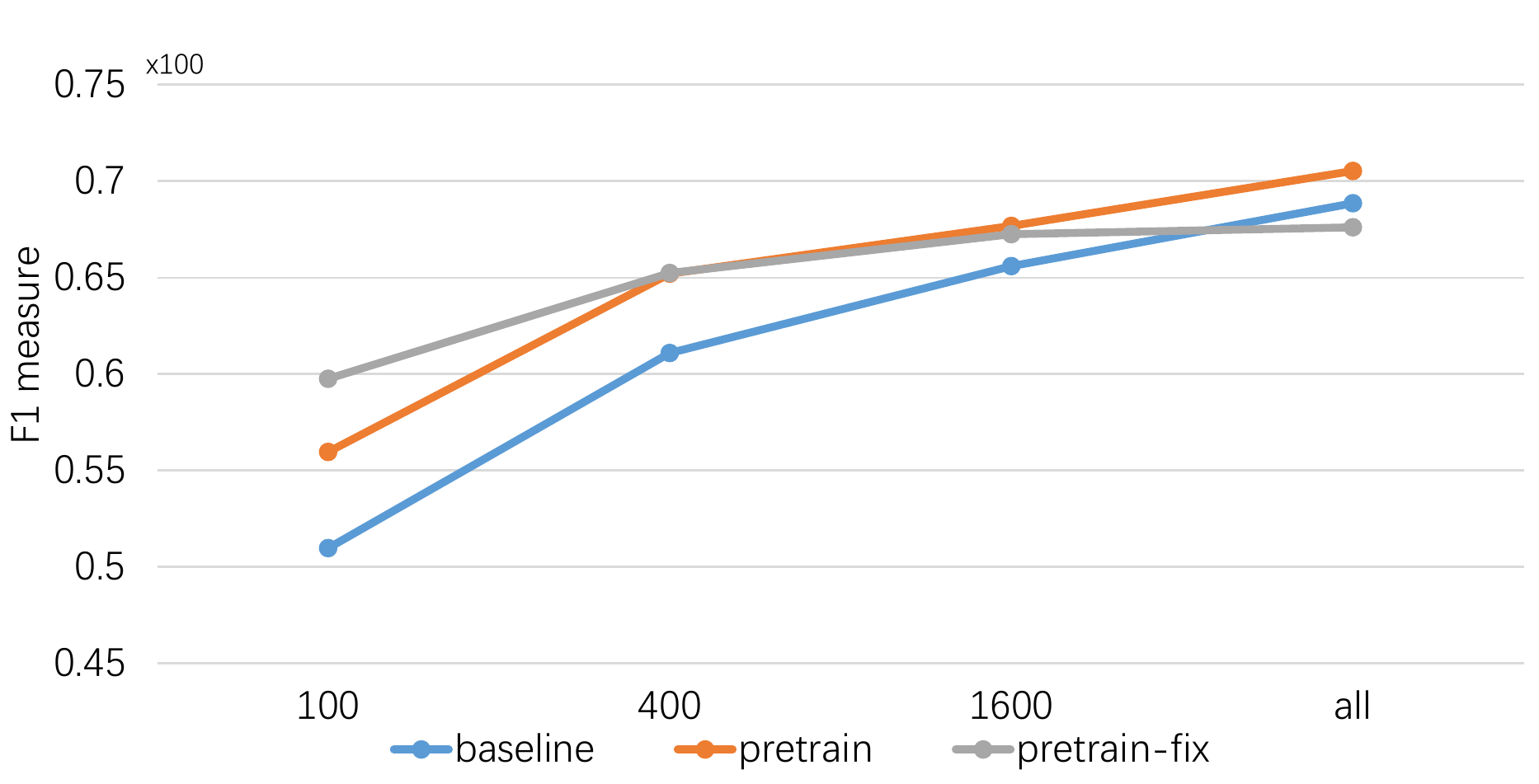}
		\caption{The mean F1 scores of different classifiers.}
		\label{fig:meanf1}
	\end{figure}

	First, we discuss the results of a few annotations. our method performs better than the random version under sub-100 and sub-400; furthermore, the mean score of pretrain-400 is near that of baseline-1600, which illustrates the potential of our method to achieve high performance fewer manual annotations. Second, from the perspective of category scores, the highest scores belong mostly to the classifiers initialized by human rules. Third, from the perspective of mean scores as shown in Fig. \ref{fig:meanf1}, the pretrained versions have higher scores than the random versions under the same manual annotations: pretrain-400 is near baseline-1600, and pretrain-1600 is near baseline-all. Another notable result is that the gap between random and pretrained initialization becomes small as the number of manual annotations increases. In conclusion, human rules help to reduce the demand for manual annotations.
	
	Two methods of parameter fine-tuning are discussed in Sect. \ref{sect:param-finetune}. Pretrain-* in Fig. \ref{fig:meanf1} indicates the first way, where the parameters of convolutional layers are updated during fine-tuning, and the pretrain-fix-* indicates that the convolutional parameters are initialized from the pretrained network and are fixed during fine-tuning. We find that the F1 scores of pretrain-fix-100 are higher than those of pretrain-100, but as the manual annotations increases, the performance of pretrain-fix trends to become stable, e.g., pretrain-fix-1600 and pretrain-fix-all have almost the same score. These results shows that the fixed fine-tuning method has better adaptability for cases with very few annotations.
	
	\section{conclusion and future work}
	In this paper, we propose a new method aimed at semantic segmentation based on 3D LiDAR data. To reduce the substantial demand for manual annotations during parameter training, we attempt to incorporate human knowledge into a neural network via parameter pretraining. To this end, we first pretrain a model with the autogenerated samples from a rule-based classifier so that human knowledge can be propagated into the network. Based on the pretrained model, only a small set of annotations are required to perform further finetuning. This method is examined extensively on a dynamic scene. The promising results indicate reduced reliance of manual annotation. Future work will consider the addition of more priors/knowledge, e.g., the spatial and temporal relationships between samples.


	\begin{IEEEbiography}[{\includegraphics[width=1in,height=1.25in,clip,keepaspectratio]{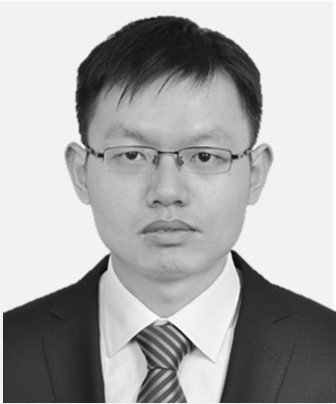}}]{Jilin Mei}
		received B.S. degree in automation in 2014 from University of Electronic Science and Technology of China, Chengdu, China. He is currently working toward the Ph.D. degree in intelligent robots in the Key Lab of Machine Perception (MOE), Peking University, Beijing, China.
		
		His research interests include intelligent vehicles, computer vision and machine learning.
	\end{IEEEbiography}
	
%	
%	\begin{IEEEbiography}[{\includegraphics[width=1in,height=1.25in,clip,keepaspectratio]{photo_gao}}]{Biao Gao}
%		received B.S. degree in computer science (machine intelligence) in 2017 from Peking University, Beijing, China. He is currently working toward the Master degree in the Key Lab of Machine Perception (MOE), Peking University, Beijing, China.
%		His research interests include intelligent vehicles,  SLAM, computer vision and machine learning.
%		
%		His research interests include intelligent vehicles, computer vision and machine learning.
%	\end{IEEEbiography}
%	
%	\begin{IEEEbiography}[{\includegraphics[width=1in,height=1.25in,clip,keepaspectratio]{photo_xudh}}]{Donghao Xu}
%		received B.S. degree in information and computing science in 2012
%		from Peking University, China. In 2018, he obtained Ph.D. degree in computer science from the same university. He is currently a post-doctoral researcher in computer science in the Key Lab of Machine Perception (MOE), Peking University, China. His research interests include computer vision, machine learning and intelligent vehicles.
%	\end{IEEEbiography}	
%	
%	\begin{IEEEbiography}[{\includegraphics[width=1in,height=1.25in,clip,keepaspectratio]{photo_yao}}]{Wen Yao}
%		Wen Yao received the Ph.D. degree in computer science (AI technology) from Peking University, Beijing, China, in 2015. He is now working with China North Vehicle Research Institute.
%		His research interests include intelligent vehicle, behavior modeling and motion planning for mobile robots.
%	\end{IEEEbiography}
%	
%	
	\begin{IEEEbiography}[{\includegraphics[width=1in,height=1.25in,clip,keepaspectratio]{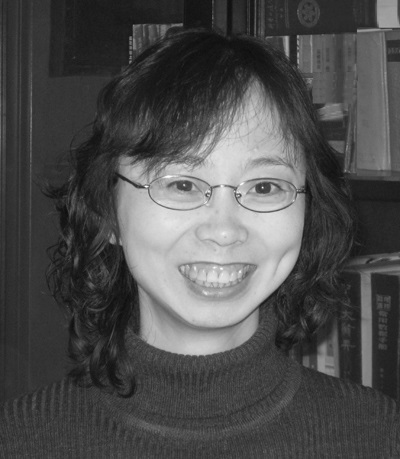}}]{Huijing Zhao}
		received B.S. degree in computer science in 1991 from
		Peking University, China. From 1991 to 1994, she was recruited by Peking
		University in a project of developing a GIS platform. She obtained M.E.
		degree in 1996 and Ph.D. degree in 1999 in civil engineering from the
		University of Tokyo, Japan. After post-doctoral research
		as the same university, in 2003, she was promoted to be a visiting
		associate professor in Center for Spatial Information Science, the
		University of Tokyo, Japan. In 2007, she joined Peking Univ as an associate
		professor at the School of Electronics Engineering and Computer Science.
		Her research interest covers intelligent vehicle, machine perception and mobile robot.
	\end{IEEEbiography}
	
\end{document}